\documentclass{article}

\usepackage[english]{babel}

\usepackage[letterpaper,top=2cm,bottom=2cm,left=3cm,right=3cm,marginparwidth=1.75cm]{geometry}

\usepackage{mathrsfs, amsmath}
\usepackage{graphicx}
\usepackage[colorlinks=true, allcolors=blue]{hyperref}
\usepackage{float}
\usepackage{listings}
\usepackage{braket}
\usepackage{color}
\usepackage{bbold}
\usepackage{tkz-tab}
\usepackage{amsmath}
\usepackage{graphicx}
\usepackage[colorlinks=true, allcolors=blue]{hyperref}
\usepackage{bbold}
\title{A multiclass Q-NLP sentiment analysis experiment using DisCoCat}
\author{%
\begin{tabular}{c} Victor Martinez \\ IBM \\ \texttt{victor.martinez@ibm.com} \end{tabular} \and
\begin{tabular}{c} Guilhaume Leroy-Meline \\ IBM \\ \texttt{guilhaume@fr.ibm.com} \end{tabular} }
\date{}
\usepackage{blindtext}
\usepackage{multicol}
\usepackage[]{natbib}

\begin{document}

\twocolumn[
\maketitle
]

\section*{Abstract}

Sentiment analysis is a branch of Natural Language Processing (NLP) which goal is to assign sentiments or emotions to particular sentences or words.
Performing this task is particularly useful for companies wishing to take into account customer feedback through chatbots or verbatim. This has been done extensively in the literature using various approaches, ranging from simple models to deep BERT neural networks. In this paper, we will tackle sentiment analysis in the Noisy Intermediate Scale Computing (NISQ) era, using the DisCoCat model of language. We will first present the basics of quantum computing and the DisCoCat model. This will enable us to define a general framework to perform NLP tasks on a quantum computer. We will then extend the two-class classification that was performed by \cite{coecke2021} to a four-class sentiment analysis experiment on a much larger dataset, showing the scalability of such a framework.

\section{Introduction to Quantum \\Computing}\label{secOne}

For a more complete introduction to Quantum Computing, the reader is invited to read \cite{nielsen00}. The base unit of any quantum computer is the \textit{qubit}, which can be seen as a two-dimensional complex vector represented by its wave function $\ket{\psi}=\alpha \ket{0}+\beta \ket{1}$. $\alpha$ and $\beta$ are complex numbers, which squared amplitude is the probability of measuring the wave function in the associated quantum state. In the same way, we may consider the system made of two uncorrelated qubits $\ket{\psi_1}=\alpha_1 \ket{0}+\beta_1 \ket{1}$ and $\ket{\psi_2}=\alpha_2 \ket{0}+\beta_2 \ket{1}$. The joint system can be written $\ket{\psi}=\alpha_1\alpha_2 \ket{00}+\alpha_1\beta_2 \ket{01}+\beta_1\alpha_2 \ket{10} + \beta_1\beta_2 \ket{11}$, or equivalently $\ket{\psi}=\ket{\psi_1}\otimes\ket{\psi_2}$, where $\otimes$ represents the tensor product. 

\begin{figure}[H]
\centering
\includegraphics[width=0.29\textwidth]{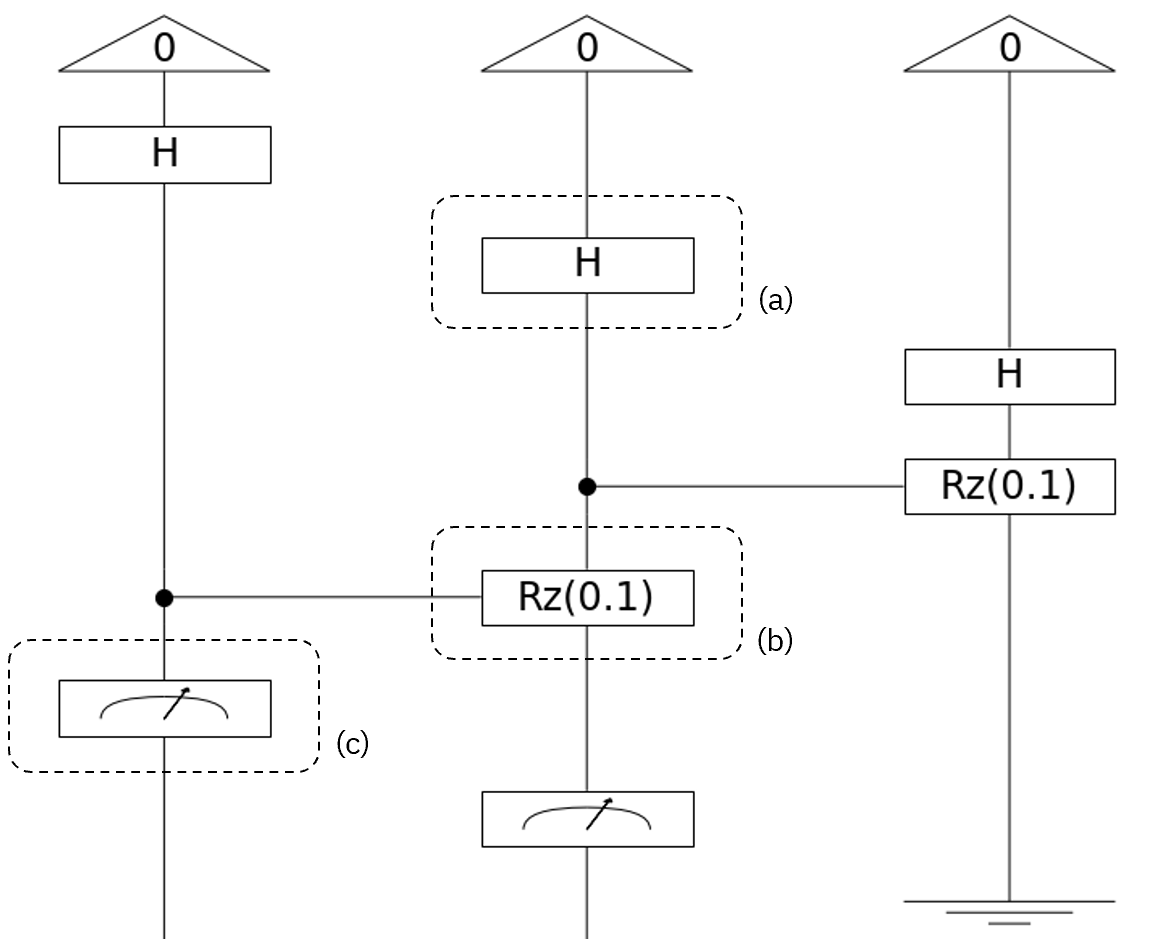}
\caption{A three qubit quantum circuit $U$ with different gates. (a) represents a Hadamard gate, (b) a parameterized controlled Z-Rotation and (c) a measurement operation}
\label{fig::bqp}
\end{figure}

Using this notation, we can generalize to $n$ uncorrelated qubits, which joint system's is written $\ket{\psi}=\ket{\psi_1}\otimes...\otimes\ket{\psi_n}$. The Born rule can be used to retrieve the probability of measuring the system $\ket{\psi}$ in state $x$ by computing $p(x)=|\braket{x|\psi}|^2$. Furthermore, one can perform elementary operations on those qubits through quantum gates. Those gates can be represented as linear algebra operations on $\ket{\psi}$ and correspond to instructions that will be physically performed on the quantum computer. The gates used along with their parameters is referred to as an \textit{Ansatz}. An example of such \textit{Ansatz} is given in Figure \ref{fig::bqp}.

This allows us to define a generic way of performing supervised machine learning tasks using variational quantum circuits. The data is first encoded in a quantum state $\ket{\psi_0}$. We then apply a parameterized quantum circuit $U(\theta)$ to $\ket{\psi_0}$, where $\theta$ is trained with the labeled dataset. We measure the wave function $\ket{\psi}=U(\theta)\ket{\psi_0}$. The result being probabilistic by nature, we perform this operation multiple times to retrieve the outcome probabilities.

\begin{figure}[H]
\centering
\includegraphics[width=0.48\textwidth]{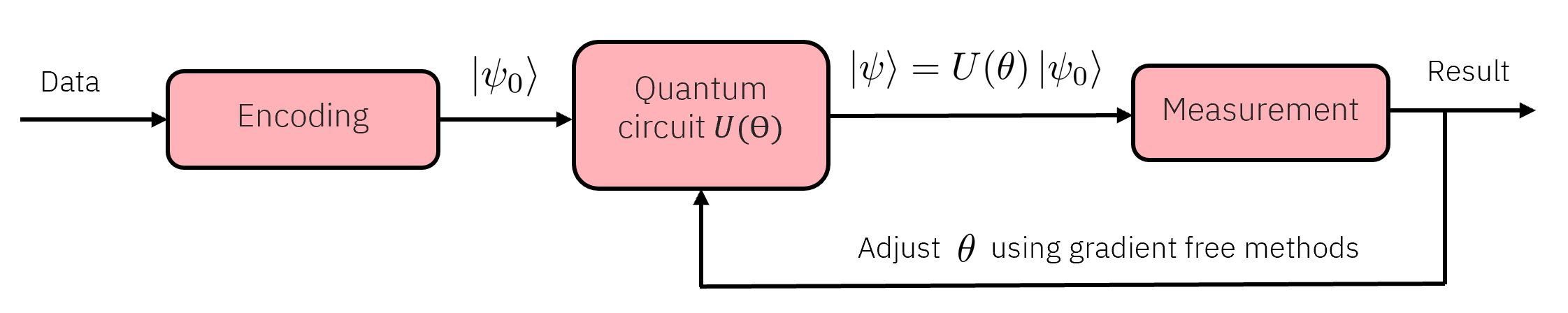}
\caption{Pipeline to perform supervised machine learning tasks on a quantum computer}
\end{figure}

The choice of \textit{Ansatz} for $U$ is crucial, since the design of the circuit has to be chosen so that the result is a function of the outcome probabilities. In some cases, the result is not directly a function of the outcome probabilities of the $n$ qubits, but a function of the outcome probabilities of $n-r$ qubits under some conditions that have to satisfy the other $r$ qubits. In Figure \ref{fig::bqp}, we could expect the second and third qubits to be $0$ for our run to be valid, the result only being given by the first qubit. \\

This requires some additional post-selection, forcing us to only keep the runs where the conditions on the $r$ qubits are met. This will be particularly important in the following, since we will expect 0 effects on many of our qubits.

\section{The DisCoCat model}\label{sectwo}

\begin{figure}[H]
\centering
\includegraphics[width=0.35\textwidth]{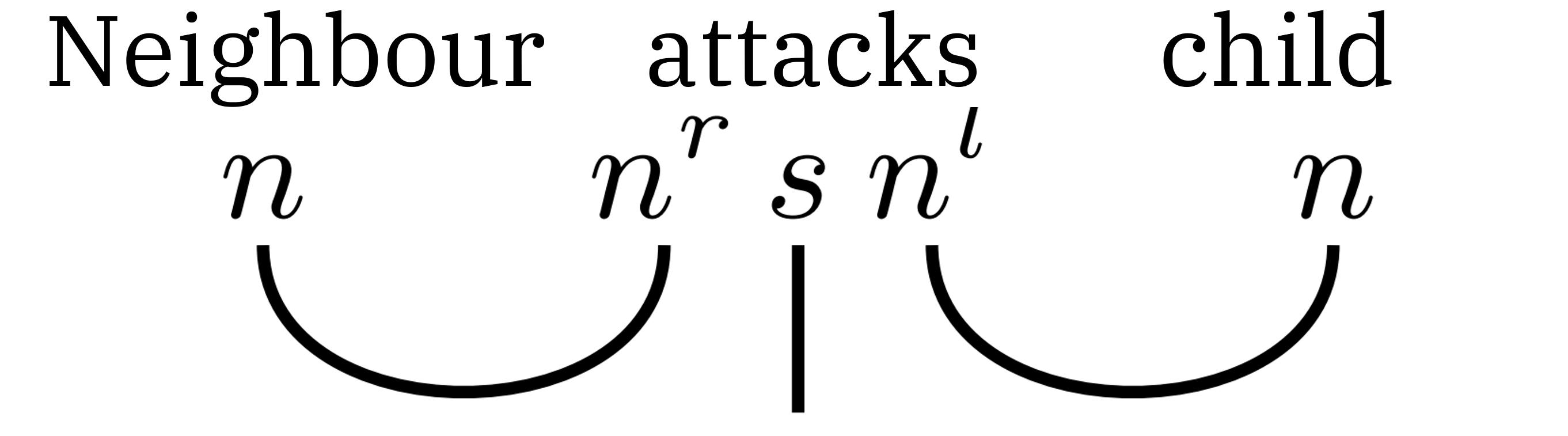}
\caption{Syntax tree of a simple noun-verb-noun sentence}
\label{fig::syntax}
\end{figure}

The categorical compositional distributional (DisCoCat) model of meaning is based on the framework of compact closed category, and the reader is invited to read \cite{coecke2010} for a complete introduction. The main idea behind DisCoCat is to decompose the language into two components. We first associate each word a meaning, as commonly done using word embedding. This word embedding can be done by projecting the words on a vector space made of basis vocabulary (for example 700 words). The second component of our language is the syntactic structure of the sentences, which is commonly represented by a syntax tree. How the words relate to each other tells us the global meaning of the sentence. An example of such syntax tree can be found in Figure \ref{fig::syntax}

Taking into account both those aspects can be done through the use of tensor products by following the below steps.

\begin{enumerate}
    \item Compose the tensor product of every word through their word embedding.\\ $\mathbf{s}=w_1 \otimes w_2 \otimes ... \otimes w_n$
    \item Construct a linear map $\mathcal{F}$ corresponding to the sentence's syntax tree.
    \item Apply the linear map to the vector $\mathbf{s}$, retrieving the meaning of the sentence $\mathcal{S}=\mathcal{F}(\mathbf{s})$.
\end{enumerate}

\begin{figure}[H]
\centering
\includegraphics[width=0.48\textwidth]{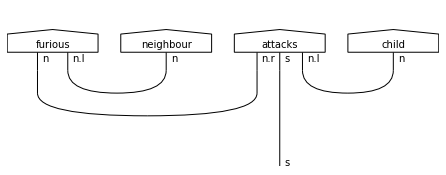}
\caption{DisCoCat diagram of the sentence Furious neighbour attacks child}
\label{fig::discocat}
\end{figure}

Why this DisCoCat model is particularly suited to quantum computing now becomes apparent. The Tensor Product Representation (TPR) requires to store the tensor product between $n$ words, which scales exponentially in $n$. This makes this model unusable on classical computers, with several terabytes required to store a single sentence. However, as seen in Section \ref{secOne}, storing the tensor product of $n$ words will scale linearly in the number of \textit{qubits}. A more convenient way of handling these TPR is through the direct use of DisCoCat diagrams. This representation is illustrated in Figure \ref{fig::discocat}, where boxes denote tensors and links tensor contractions.

\noindent Using this quantum-suited representation of language as our data encoding process, we can take a closer look on how to perform NLP tasks.

\section{General pipeline for QNLP tasks}

We can define a general framework for performing QNLP tasks with the steps below. 

\begin{enumerate}
    \item The sentence is parsed to obtain its syntax tree. For large datasets, this would require an automatic parser.
    \item The syntax tree and the sentence are then combined to obtain the DisCoCat diagram which is written in an efficient way.
    \item The DisCoCat diagram is mapped to a parameterized quantum circuit, which parameters are obtained through training.
    \item The quantum circuit is compiled and executed on a quantum computer multiple times to retrieve the measurement statistics.
    \item Post-processing is performed to select only runs where the $r$ qubits have 0 effects. This gives us the final result.
\end{enumerate}

\begin{figure}[H]
\centering
\includegraphics[width=0.48\textwidth]{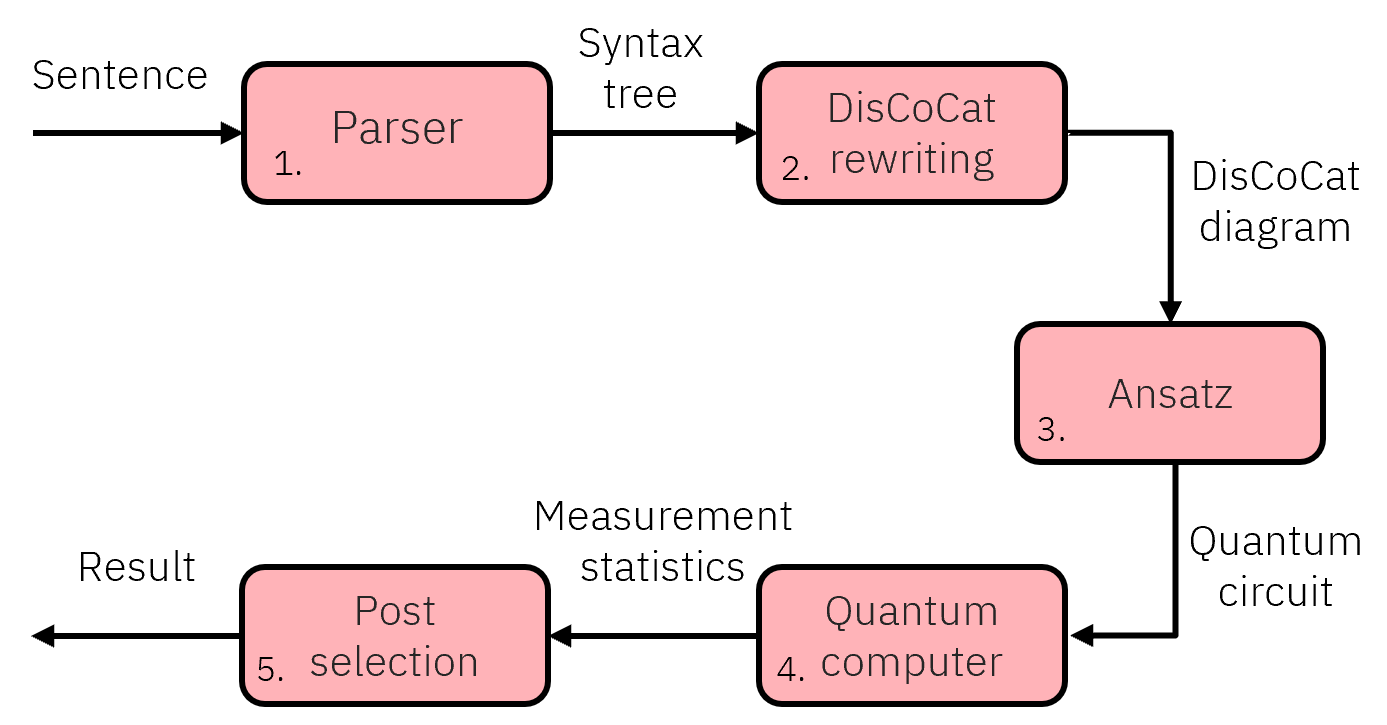}
\caption{Pipeline to perform NLP tasks on a quantum computer}
\label{fig::nlppiple}
\end{figure}

\noindent Those steps are summarised in Figure \ref{fig::nlppiple}. In Section \ref{sectquatre}, those steps will be detailed on a real sentiment analysis task.

\section{Sentiment analysis experiment}\label{sectquatre}

\cite{coecke2021} defined this framework and tested it on two-class meaning classification, yielding promising results. In this paper, we propose a similar approach to perform sentiment analysis on four emotions with a larger dataset. We restrict ourselves to the four most basic emotions: \textit{happiness, fear, anger, sadness}. We will be using 3 and 4 words sentences, since we will be working with simulators or small size quantum computers. \\

To circumvent the absence of automatic parser at our disposition, we create the sentences based on their syntax. We first start by defining the nouns, adjectives, transitive and intransitive verbs that will constitute our dataset. For each syntax - for example adjective-noun-verb-noun - we create every sentence possible matching this syntax. We therefore have access to the syntax of every sentence in our dataset.

\begin{table}[H]
\centering
 \begin{tabular}{ |p{2cm}|p{2cm}|p{2cm}|}
 \hline
 Nouns & Adjectives & Verbs \\
 \hline
& anxious & attack\\
neighbour  & ecstatic  & scare\\
   & irritated & anger\\
   & distressed & amuse\\
 child & blissful & demoralise\\
   & furious & cry \\
   & petrified & laugh \\
 boy & frightened & dance \\
 & miserable & scream \\
  & young, blind  &\\
  & cheerful&\\
 \hline
\end{tabular}
\end{table}

\begin{figure}[H]
\centering
\includegraphics[width=0.44\textwidth]{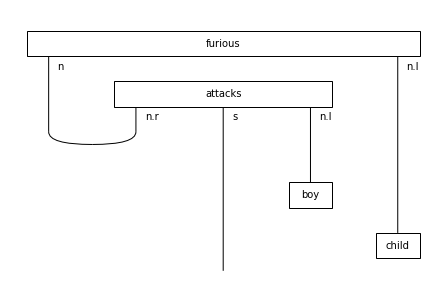}
\includegraphics[width=0.44\textwidth]{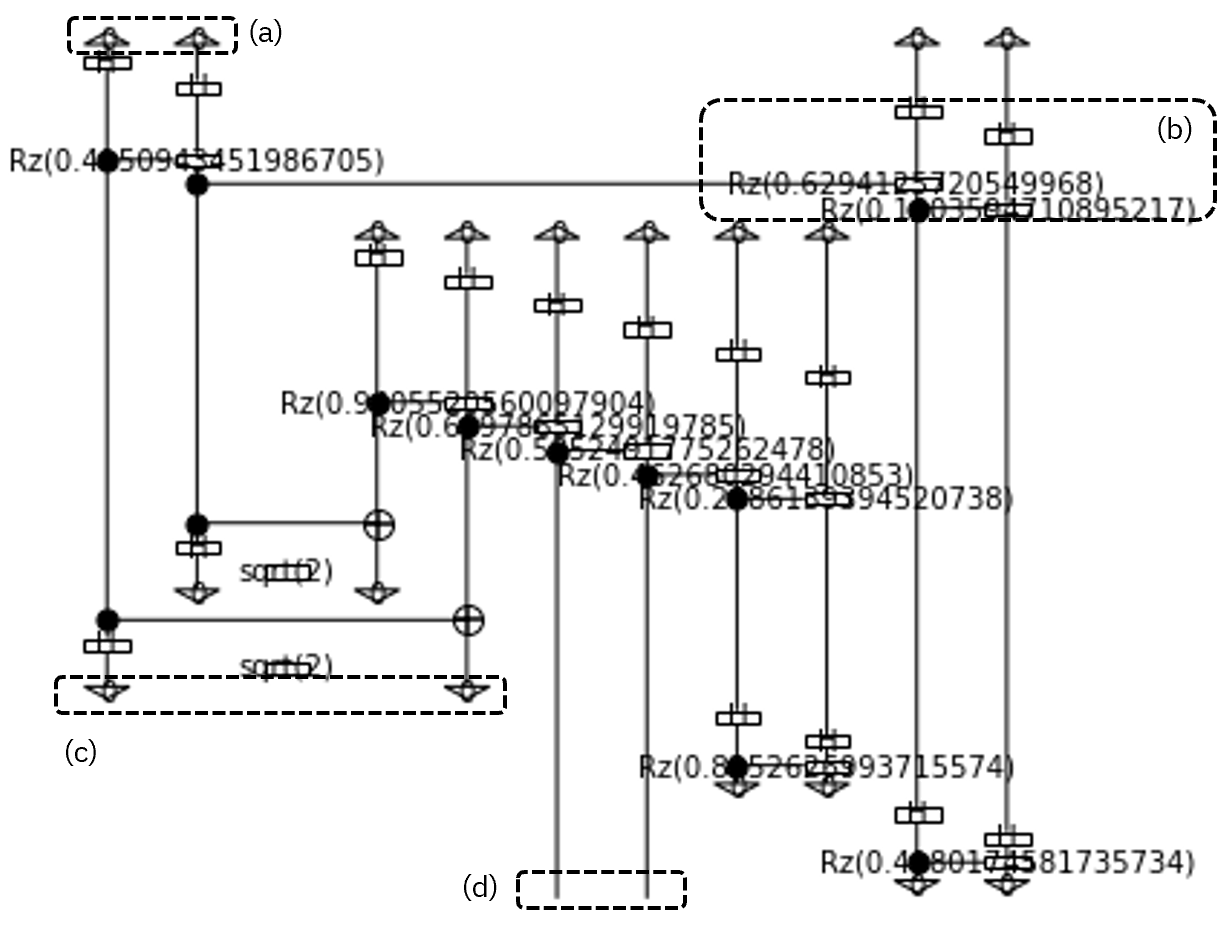}
\caption{DisCoCat diagram and its associated quantum circuit. (a) Top qubits represent the initial qubits. (b) Gates applied to the qubits. (c) Bottom qubits expected to be 0. (d) Measured qubits giving the result probabilities}
\label{fig::qcdisco}
\end{figure}

The second step, corresponding to the writing of the DisCoCat diagram, is done through DiscoPy, an  open source toolbox for computing with monoidal categories developed by \cite{de_Felice_2021}. This allows us to efficiently write the diagrams and map them to quantum circuits. An example of a DisCoCat diagram and its associated quantum circuit is depicted on Figure \ref{fig::qcdisco}. The parametrisation of the \textit{Ansatz} was done following \cite{coecke2021} and led to using two qubits for verbs and nouns.

\begin{figure}[H]
\centering
\includegraphics[width=0.42\textwidth]{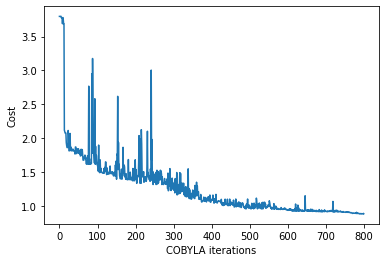}
\caption{Evolution of the cross entropy on the training dataset}
\label{fig::costfunctoin}
\end{figure}

We run the circuits multiple time with post-selection and measure the two qubits of interest to retrieve the outcome probabilities ($\ket{00}$, $\ket{01}$, $\ket{10}$, $\ket{11}$). These outcome probabilities directly give us the emotion linked to our sentence.  The parameters of the \textit{Ansatz} are trained using a Cross-entropy loss and the outcome probabilities. The optimiser chosen was COBYLA, a gradient free optimizer. We split our dataset into a training and test dataset and plot the evolution of the cost function on the training dataset in Figure \ref{fig::costfunctoin}.

Once the training of our variational quantum circuits has converged, we measure our accuracy on the test dataset made of 180 sentences. We plot the results in the form of a normalized confusion matrix, comparing the ground truth and prediction. 

\begin{figure}[H]
\centering
\includegraphics[width=0.45\textwidth]{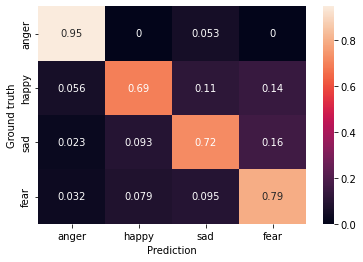}
\caption{Normalized confusion matrix on the test dataset}
\label{fig::conf}
\end{figure}
 
We reach satisfying accuracy results with an F1-score of $78\%$, with a confusion matrix concentrated on the diagonal. 

\section{Conclusion and prospects}\label{conc}

The DisCoCat model of meaning, by its structure and the computations it entails, seems to be an adequate way of tackling NLP problems on quantum machines. The potential scalability of the presented experiment bodes well for the future. The sentence length, the number of sentences and the vocabulary basis all grow linearly or sublinearly in the number of qubits.
Running independent evaluations of quantum circuits on multiple quantum processors -if they become more accessible-, could be an adequate way of handling huge amount of sentences. With quantum processors of around 1000 qubits, performing NLP experiments on whole texts could
be possible. In this paper, we have started to show the potential scalability of this method, realising for the first time a sentiment analysis with four emotions with a dataset of over 800 sentences.\\

However, it is still important to remind that we are to this day still far from a quantum advantage. For the experiment presented in this paper, a simple TF-IDF feature extractor coupled with a Naive Bayes classifier reached $95\%$ accuracy. This can be explained by the simplicity of the sentences used along with the small depth of the quantum circuits at hand. In a more general way, supervised learning can be directly linked to kernel methods, with the mapping to the higher dimensional Hilbert space corresponding to the encoding of the data on the quantum computer. For a quantum advantage to arise, the kernel linked to this higher dimensional mapping has to be hard to simulate classically, as shown by \cite{mariaschuld}.

\bibliographystyle{apalike}
\bibliography{Syn3Ref}

\end{document}